\documentclass{article}

\usepackage[final]{neurips_2021}

\usepackage[utf8]{inputenc} 
\usepackage[T1]{fontenc}    
\usepackage{hyperref}       
\usepackage{url}            
\usepackage{booktabs}       
\usepackage{amsfonts}       
\usepackage{nicefrac}       
\usepackage{microtype}      
\usepackage{xcolor}         
\usepackage{graphicx}
\usepackage{enumitem}
\usepackage{algorithm}
\usepackage{algorithmic}
\usepackage{xspace}
\usepackage{wrapfig}
\usepackage[export]{adjustbox}
\usepackage{capt-of}
\newcommand{\ar}{\texttt{AR-AdaLS}\xspace}
\newcommand{\arensem}{\texttt{AR-AdaLS of Ensemble}\xspace}
\newcommand{\ensemar}{\texttt{Ensemble of AR-AdaLS}\xspace}
\newcommand{\ada}{\texttt{AdaLS}\xspace}
\title{Improving Calibration through the Relationship with Adversarial Robustness}

\author{%
Yao Qin \hspace{0.6cm} Xuezhi Wang \hspace{0.6cm} Alex Beutel \hspace{0.6cm} Ed H. Chi\\
Google Research\\
\texttt{ \{yaoqin, xuezhiw, alexbeutel, edchi\}@google.com }
}

\begin{document}

\maketitle

\begin{abstract}
Neural networks lack \emph{adversarial robustness}, i.e., they are vulnerable to adversarial examples that through small perturbations to inputs cause incorrect predictions. Further, trust is undermined when models give \emph{miscalibrated} predictions, i.e.,  the predicted probability is not a good indicator of how much we should trust our model. In this paper, we study the connection between adversarial robustness and calibration and find that the inputs for which the model is sensitive to small perturbations (are easily attacked) are more likely to have poorly calibrated predictions. Based on this insight, we examine if calibration can be improved by addressing those adversarially unrobust inputs. To this end, we propose Adversarial Robustness based Adaptive Label Smoothing (\ar) that integrates the correlations of adversarial robustness and calibration into training by adaptively softening labels for an example based on how easily it can be attacked by an adversary. We find that our method, taking the adversarial robustness of the in-distribution data into consideration, leads to better calibration over the model even under distributional shifts. In addition, \ar~can also be applied to an ensemble model to further improve model calibration. 
\end{abstract}

\section{Introduction}
    The robustness of machine learning algorithms is becoming increasingly important as ML systems are being used in higher-stakes applications. 
    In one line of research, neural networks are shown to lack \emph{adversarial robustness} -- small perturbations to the input can successfully fool classifiers into making incorrect predictions~\citep{szegedy2013, Goodfellow2014ExplainingAH, carlini2017towards, madry2017towards, qin2019detecting}. In largely separate lines of work, researchers have studied uncertainty in model's predictions.  For example, models are often \emph{miscalibrated} where the predicted confidence is not indicative of the true likelihood of the model being correct \citep{Guo2017OnCO, mixup, ensemble, batch_ensemble, class_calibration}.  
    The calibration issue is exacerbated when models are asked to make predictions on data different from the training distribution~\citep{snoek2019can}, which becomes an issue in practical settings where it is important that we can trust model predictions under distributional shift.
    
    Despite robustness, in all its forms, being a popular area of research,
    the \emph{relationship} between these perspectives has not been extensively explored previously. 
In this paper, we study the correlation between adversarial robustness and calibration. 
    We discover that input data points that are sensitive to small adversarial perturbations (are easily attacked) are more likely to have poorly calibrated predictions. This holds true on a number of network architectures for classification and on all the datasets that we consider: CIFAR-10~\citep{krizhevsky2009learning}, CIFAR-100~\citep{krizhevsky2009learning} and ImageNet~\citep{Russakovsky2015ImageNetLS}. This suggests that the miscalibrated predictions on those adversarially unrobust data points greatly degrades the performance of model calibration. Based on this insight, we hypothesize and study if calibration can be improved by giving different supervision to the model depending on adversarial robustness of each training data.
    
    To this end, we propose \textbf{A}dversarial \textbf{R}obustness based \textbf{Ada}ptive \textbf{L}abel \textbf{S}moothing (\ar) to integrate the correlations between adversarial robustness and calibration into training. Specifically, \ar adaptively smooths the training labels conditioned on how vulnerable an input is to adversarial attacks. Our method improves label smoothing~\citep{szegedy2013} by explicitly teaching the model to differentiate the training data according to their adversarial robustness and then adaptively smooth their labels. By giving different supervision to the training data, our method leads to better calibration over the model without an increase of latency during inference.
    In addition, since adversarially unrobust data points can be considered as outliers of the underlying data distribution~\citep{Carlini2019DistributionDT}, our method can even greatly improve model calibration on held-out shifted data. Further, we propose ``\arensem'' to combine our \ar and deep ensembles~\citep{ensemble, snoek2019can}, to further improve the calibration performance under distributional shift.  Last, we find an additional benefit of \ar is improving model stability (i.e., decreasing variance over multiple independent runs), which is valuable in practical applications where changes in predictions across runs (churn) is problematic.
    
In summary, our main contributions are as follows:
\begin{itemize}[leftmargin=20pt]
    \item \textbf{Relationship among Robustness Metrics:} 
    We find a significant correlation between adversarial robustness and calibration: inputs that are unrobust to adversarial attacks are more likely to have poorly calibrated predictions.
    \item \textbf{Algorithm:} 
    We hypothesize that training a model with different supervision based on adversarial robustness of each input will make the model better calibrated. To this end, we propose \ar to automatically learn how much to soften the labels of training data based on their adversarial robustness.
    Further, we introduce ``\arensem'' to show how to apply \ar to an ensemble model.
    \item \textbf{Experimental Analysis:} On CIFAR-10, CIFAR-100 and ImageNet, we find that 
    \ar is more effective than previous label smoothing methods in improving calibration, particularly for shifted data. Further, we find that while ensembling can be beneficial, applying \ar to adaptively calibrate ensembles offers further improvements over calibration. 
\end{itemize}

\begin{figure*}[t]
     \centering
      \includegraphics[width=0.3\linewidth]{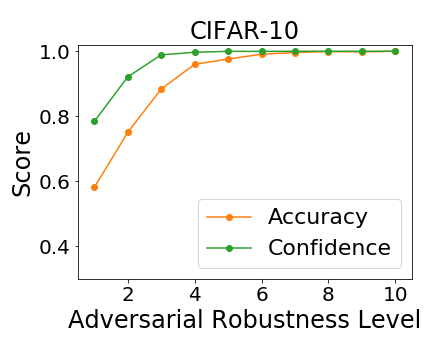}
      \includegraphics[width=0.3\linewidth]{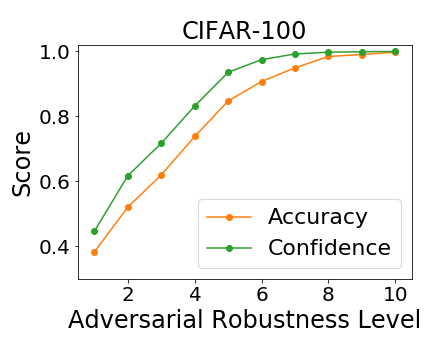}
      \includegraphics[width=0.3\linewidth]{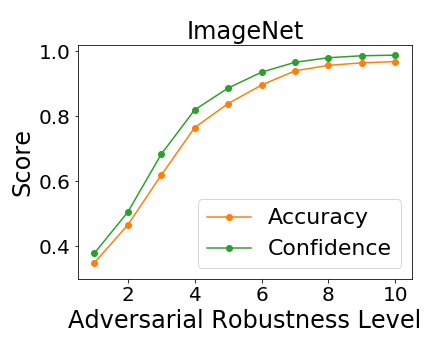}\\
    \vspace{-3mm}
      \includegraphics[width=0.3\linewidth]{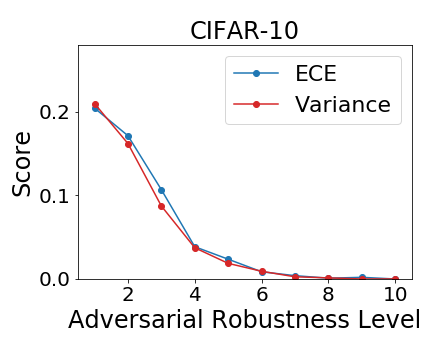}
\includegraphics[width=0.3\linewidth]{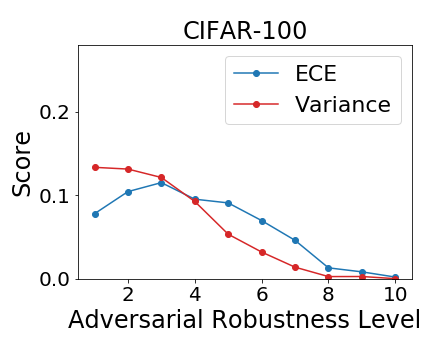}
      \includegraphics[width=0.3\linewidth]{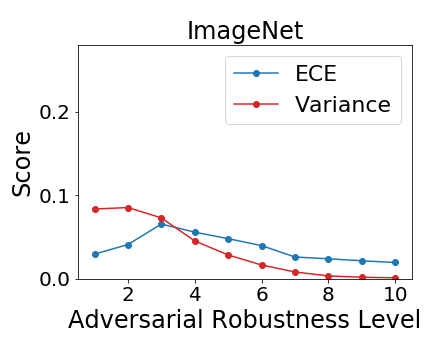}
      \vspace{-2mm}
     \caption{Inputs that are adversarially unrobust are more likely to have poorly calibrated and unstable predictions on CIFAR-10, CIFAR-100 and ImageNet. \textbf{Top:} Accuracy and confidence of the predicted class. \textbf{Bottom:} ECE (lower is better) and variance (lower is better) in each adversarial robustness subset. Higher adversarial robustness level means the input are more adversarially robust (require larger adversarial perturbations to fool the classifier into wrong predictions).}
  \label{fig:cor}
  \vspace{-3mm}
\end{figure*}  
\section{Related Work}
 \paragraph{Uncertainty estimates}
 How to better estimate a model's predictive uncertainty is an important research topic, since many models with a focus on accuracy may fall short in predictive uncertainty. A popular way to improve a model's predictive uncertainty is to make the model well-calibrated, e.g., post-hoc calibration by temperature scaling \citep{Guo2017OnCO}, and multi-class Dirichlet calibration \citep{class_calibration}.
 In addition, Bayesian neural networks, through learning a posterior distribution over network parameters, can also be used to quantify a model's predictive uncertainty, e.g., \citet{NIPS2011_4329, pmlr-v37-blundell15, mcmc}. Dropout-based variational inference~\citep{gal2016dropout, kingma2015variational} can help DNN models make less over-confident predictions and be better calibrated. Recently, mixup training \citep{Zhang2018mixupBE} has been shown to improve both models' generalization and calibration \citep{mixup}, by preventing the model from being over-confident in its predictions. 
Despite the success of improving uncertainty estimates over in-distribution data, \citet{snoek2019can} argue that it does not usually translate to a better performance on data that shift from the training distribution. Among all the methods evaluated by \citet{snoek2019can} under distributional shift, ensemble of deep neural networks \citep{ensemble}, is shown to be most robust to dataset shift, producing the best uncertainty estimates. 
\paragraph{Adversarial robustness}
 On the other hand, machine learning models are known to be  brittle \citep{xin2017folding} and  vulnerable to adversarial examples~\citep{anish2018, carlini2017_adv, carlini2017towards, warren2018, qin2020deflecting}. Many defenses have been proposed to improve model's adversarial robustness~\citep{song2017pixeldefend, yang2019me, goodfellow2018evaluation}, however are further attacked by more advanced defense-aware attacks~\citep{carlini2017towards, anish2018}. Recently, ~\citet{Carlini2019DistributionDT, Stock2018ConvNetsAI} define adversarial robustness as the minimum distance in the input domain required to change the model's output prediction by constructing an adversarial attack. The most recent work that is close to ours, \citet{Carlini2019DistributionDT}, makes the interesting observation that easily attackable data are often outliers in the underlying data distribution and then use adversarial robustness to determine an improved ordering for curriculum learning.
  Our work, instead, explores the relationship between adversarial robustness and calibration. In addition, we use adversarial robustness as an indicator to adaptively smooth the training labels to improve model calibration.
\paragraph{Label smoothing} Label smoothing is originally proposed in~\citet{Szegedy2016RethinkingTI} and is shown to be effective in improving the quality of uncertainty estimates in~\citet{Mller2019WhenDL,mixup}. Instead of  minimizing the cross-entropy loss between the predicted probability $\hat{p}$ and the one-hot label $p$, label smoothing minimizes the cross-entropy between the predicted probability and a softened label $\widetilde p = p (1 - \epsilon) + \frac{\epsilon}{Z}$, where $Z$ is the number of classes in the dataset and $\epsilon$ is a hyperparameter 
which controls the degree of the smoothing effect. Our work makes label smoothing adaptive and incorporates the correlation with adversarial robustness to further improve calibration.

\section{Correlations between Adversarial Robustness and Calibration}\label{sec:cor}
To explore the relationship between adversarial robustness and calibration, we first introduce the metrics to evaluate each of them (arrows indicate which direction is better).
\paragraph{Adversarial robustness $\uparrow$} Adversarial robustness measures the minimum distance in the input domain required to change the model's output prediction by constructing an adversarial attack~\citep{Carlini2019DistributionDT, Stock2018ConvNetsAI}. Specifically, given an input $x$ and a classifier $f(\cdot)$ that predicts the class for the input, the adversarial robustness is defined as the minimum adversarial perturbation $\delta$ that enables $f(x + \delta) \neq f(x)$. Following the work~\citep{Carlini2019DistributionDT}, we construct the $\ell_2$ based CW attack~\citep{carlini2017towards} and then use the $\ell_2$ norm of the adversarial perturbation $\|\delta \|_2$ to measure the distance to the decision boundary. Therefore, a more adversarially robust input requires a larger adversarial perturbation to change the model's prediction.
\paragraph{Expected calibration error $\downarrow$} Model calibration measures the alignment between the predicted probability and the accuracy. Well calibrated predictions convey the information about how much we should trust a model's prediction. We follow the widely used expected calibration error (\textbf{ECE}) to measure the calibration performance of a network~\citep{Guo2017OnCO, snoek2019can}. To compute the ECE, we need to first divide all the data into $K$ buckets sorted by their predicted probability (confidence) of the predicted class. Let $B_k$ represent the set of data in the $k$-th confidence bucket. Then the accuracy and the confidence of $B_k$ are defined as $\mathrm{acc}(B_k) = \frac{1}{|B_k|}\sum_{i\in B_k} {\bf{1}}(\hat{y}_i = y_i)$ and $\mathrm{conf}(B_k) = \frac{1}{|B_k|}\sum_{i\in B_k} \hat{p}^{\hat{y}_i}_i$, where $\hat{y}$ and $y$ represent the predicted class and the true class respectively, and $\hat{p}^{\hat{y}}$ is the predicted probability of $\hat{y}$. The ECE is then defined as $\mathrm{ECE} = \sum_{k=1}^K \frac{|B_k|}{N} |\mathrm{acc}(B_k) - \mathrm{conf}(B_k)|$,
where $N$ is the number of data points.

\begin{table*}[!t]
\centering
\vspace{-5mm}
\caption{Network architecture and accuracy used for each dataset.}
\label{tab:net}
\small
\begin{tabular*}{\textwidth}{c @{\extracolsep{\fill}} ccc}
\toprule
Dataset    & CIFAR-10     & CIFAR-100            & ImageNet      \\
\midrule
Network   & ResNet-29 & WRN-28-10  & ResNet-101\\
Accuracy & 91.4\%       & 79.2\%               &      77.7\%         \\
\bottomrule            
\end{tabular*}
\vspace{-4mm}
\end{table*}
\subsection{Correlations}
Based on the evaluation metrics, we can see that adversarial robustness and calibration are measuring quite different properties: the adversarial robustness measures the property of the data by computing the adversarial perturbation $\delta$ from the \emph{input domain}, while the calibration metric measures the properties of the model's predicted probability in the \emph{output space}. Although adversarial robustness and calibration are conceptually different, they are both connected to the decision boundary. Specifically, adversarial robustness can be used to measure the distance to the decision boundary: if a data point is adversarially unrobust, i.e., easy to find a small input perturbation to fool the classifier into wrong classifications, then this data point is close to the decision boundary. Meanwhile, models should have relatively less confident predictions on data points close to the decision boundary. However, as pointed out by~\citep{Guo2017OnCO, snoek2019can}, existing deep neural networks are frequently over-confident, i.e., having predictions with high confidence even whey they should be uncertain. Taking these two together, we hypothesize if \emph{examples that can be easily attacked by adversarial examples are also poorly calibrated}.

To test this, we perform experiments on the clean test set across three datasets: CIFAR-10~\citep{krizhevsky2009learning}, CIFAR-100~\citep{krizhevsky2009learning} and ImageNet~\citep{Russakovsky2015ImageNetLS} with different networks, whose architecture and accuracy are shown in Table~\ref{tab:net}. We refer to these models as ``Vanilla'' for each dataset in the following discussion. 
The details for training each vanilla network are included in Appendix A.

To explore the relationship between adversarial robustness and calibration, we start with the relationship between adversarial robustness and confidence together with accuracy. Specifically, we rank the input data according to their adversarial robustness and then divide the dataset into $R$ equally-sized subsets ($R=10$ used in this paper). For each adversarial robustness subset, we compute the accuracy and the average confidence score of the predicted class. As shown in the first row in Figure~\ref{fig:cor}, we can clearly see that both accuracy and confidence increase with the adversarial robustness of the input data, and confidence is consistently higher than accuracy in each adversarial robustness subset across three datasets. This indicates that although vanilla classification models achieve the state-of-the-art accuracy, they tend to give over-confident predictions, especially for those adversarially unrobust data points.

Taking one step further, we particularly compute the expected calibration error (ECE) in each adversarial robustness subset, shown in the bottom row of Figure~\ref{fig:cor}.  
In general, we find that data points falling into lower adversarial robustness levels are more likely to be over-confident and less well calibrated (larger ECE). For those adversarially robust examples, there is a better alignment between the model's predicted confidence and accuracy, and the ECE over those examples is close to 0. This nicely validates our hypothesis: inputs that are adversarially unrobust are more likely to have poorly calibrated predictions.
On larger-scale ImageNet, while we still see the general trend holds, the least adversarially robust examples are relatively well calibrated. We hypothesize this may be due to larger training data and less overfitting.

Furthermore, we also find an interesting correlation between adversarial robustness and model stability, which is measured by the variance of the predicted probability across $M$ independent runs (e.g., $M=5$). The variance is computed as $\sigma^2 = \frac{1}{M-1}\frac{1}{N}\sum_{m=1}^M\sum_{i=1}^N (\hat{p}_{m,i} - \bar{p}_{i})^2$,
where $\hat{p}_{m,i}$ is the $m$-th model's predicted probability of the $i$-th data and $\bar{p}_i = \frac{1}{M}\sum_{m=1}^M \hat{p}_{m,i}$ is the average predicted probability over $M$ runs.
As shown in the bottom row of Figure~\ref{fig:cor}, we see that those adversarially unrobust examples tend to have a much higher variance across all three datasets. This indicates that inputs that are unrobust to adversarial attacks are more likely to have unstable predictions.

Taking all together, these empirical results nicely build a connection between very different concepts. In particular, adversarial robustness is measured over the input domain while both calibration and stability are measured over the output space. Given the strong empirical connection, we now ask: \emph{can we improve model calibration and stability by targeting adversarially unrobust examples?}
\begin{algorithm}[tb]
  \caption{Training procedure for \ar}
  \label{im}
\begin{algorithmic}
\STATE \textbf{Input:} number of classes $Z$, number of training epochs $T$, number of adversarial robustness subset $R$, learning rate of adaptive label smoothing $\alpha$. 
\STATE For each adversarial robustness training subset, we initialize the soft label as the one-hot label $\widetilde{p}_{r, t}=p_r$, 
where the initial soft label for the correct class $\widetilde{p}_{r, t}^{z=y}=1$.
  \FOR{$t=1$ {\bfseries to} $T$}
  \STATE Minimize cross-entropy loss between soft label and predicted probability $\frac{1}{R}\sum_r^R \mathcal{L}(\widetilde{p}_{r,t}, \hat{p}_{r,t})$
  \FOR{$r=1$ {\bfseries to} $R$}
  \STATE Update $\widetilde{p}^{z=y}_{r, t+1} \leftarrow \widetilde{p}^{z=y}_{r, t} - \alpha \cdot \{\mathrm{conf}(S^{val}_r)_t - \mathrm{acc}(S^{val}_r)_t\}$  \hfill $\triangleright$ Eqn. (\ref{eqn:softlabel})
  \STATE Clip $\widetilde{p}^{z=y}_{r, t+1}$ to be within ($\frac{1}{Z}, 1$]
  \STATE Update $\epsilon_{r, t+1} \leftarrow (\widetilde{p}^{z=y}_{r, t+1} - 1) \cdot \frac{Z}{1 - Z}$ \hfill $\triangleright$ Eqn. (\ref{eqn:epsilon})
  \STATE Update $\widetilde p_{r,t+1} \leftarrow p_r (1 - \epsilon_{r, t+1}) + \frac{\epsilon_{r,t+1}}{Z}$ \hfill $\triangleright$ Eqn. (\ref{eqn:y})
  \ENDFOR
  \ENDFOR
\end{algorithmic}
\end{algorithm}

\section{Method}\label{sec:alg}
Based on the correlation between adversarial robustness and calibration, we hypothesize and study if calibration can be improved by giving different supervision to the model depending on the adversarial robustness of training data. 
To this end, we propose a method named \textbf{A}dversarial \textbf{R}obustness based \textbf{Ada}ptive \textbf{L}abel \textbf{S}moothing (\textbf{\ar}), which performs
label smoothing at different degrees to the training data based on their adversarial robustness. Specifically, we sort and divide the training data into $R$ small subsets with equal size according to their adversarial robustness\footnote{Note, predicted confidence is not a good indicator for splitting the training dataset as the model can easily overfit to the training data and their predicted confidence are all close to 100$\%$.}  and then use $\epsilon_r$ to soften the labels in each training subset $S_r^{train}$. The soft labels can be formulated as:
\begin{equation}\label{eqn:y}
    \widetilde p_r = p_r (1 - \epsilon_r) + \frac{\epsilon_r}{Z},
\end{equation} 
where $p_r$ stands for the one-hot vector, e.g., $p^{z=y}_r=1$ for the correct class and $p^{ z \neq y}_r=0$ for the others, and $Z$ is the number of classes in the dataset. The parameter $\epsilon_r$ controls the degree of smoothing effect and allows for different levels of smoothing in each adversarial robustness subset.
Generally, a relatively larger $\epsilon_r$ is desirable for lower adversarial robustness levels such that the model learns to make a lower confident prediction. 
Instead of empirically setting the parameter $\epsilon_r$ in each adversarial robustness subset, we allow it to be adaptively updated according to the calibration performance on the validation set (discussed in Section~\ref{sec:adaptive}). In this way, we explicitly train a network with different supervision based on the adversarial robustness of training data. 

There are two options to obtain the adversarial robustness. One is ``on the fly'': to keep creating the adversarial attacks during training, which provides precise adversarial robustness ranking but at the cost of great computing time. The other is to ``pre-compute'' the adversarial robustness by attacking a vanilla model with the same network architecture but trained with one-hot labels. This is more efficient but at the sacrifice of the precision of adversarial robustness ranking. In practice, we find that it is sufficient to make the network differentiate the adversarially robust and unrobust data with the pre-computed adversarial robustness (see more discussion in Section~\ref{sec:ablation}). Therefore, all experiments related to ``\ar'' without further specification are based on pre-computed adversarial robustness for efficiency.

\subsection{Adaptive learning mechanism}\label{sec:adaptive} 
To find the best hyperparameter $\epsilon$ for label smoothing, previous methods~\citep{Szegedy2016RethinkingTI, mixup} sweep $\epsilon$ in a range and choose the one that has the best validation performance. However, in our setting, the number of combinations of $\epsilon_r$ increases exponentially with the number of adversarial robustness subsets $R$. To this end, we propose an adaptive learning mechanism to automatically learn the parameter $\epsilon_r$ in each adversarial robustness subset. The overall training procedure is summarized in Algorithm~\ref{im}.

First, we denote the soft label for the correct class in the $r$-th adversarial robustness subset as $\widetilde{p}^{z=y}_r$. According to Eqn. (\ref{eqn:y}), we can derive:
\begin{equation}\label{eqn:ysoft}
    \widetilde{p}^{z=y}_r  = 1 - \epsilon_r + \frac{\epsilon_r}{Z}.
\end{equation}
 Since well-calibrated predicted probability should be aligned with the empirical accuracy, we use the calibration performance in the \emph{validation set} to help update $\widetilde{p}^{z=y}_r$ for the \emph{training data}. Specifically, we first rank the adversarial robustness of the validation data and split the validation set into $R$ equally-sized subsets. Then, we use the difference between confidence and accuracy in the $r$-th adversarial robustness validation subset $\mathrm{conf}(S^{val}_r) - \mathrm{acc}(S^{val}_r)$ to update the soft label for the correct class of training data in the $r$-th adversarial robustness training subset $S^{train}_r$,
\begin{equation}\label{eqn:softlabel}
  \widetilde{p}^{z=y}_{r, t+1} = \widetilde{p}^{z=y}_{r, t} - \alpha \cdot \{\mathrm{conf}(S^{val}_r)_t - \mathrm{acc}(S^{val}_r)_t\},
\end{equation}
where $\widetilde{p}^{z=y}_{r, t}$ is the soft label of the correct class in the $r$-th adversarial robustness training subset at time step $t$. The accuracy and the confidence of $S^{val}_r$ are defined as $\mathrm{acc}(S^{val}_r) = \frac{1}{|S^{val}_r|}\sum_{i\in S^{val}_r} {\bf{1}}(\hat{y}_i = y_i)$ and $\mathrm{conf}(S^{val}_r) = \frac{1}{|S^{val}_r|}\sum_{i\in S^{val}_r} \hat{p}^{z=\hat{y}_i}_i$, where $\hat{y}$ and $y$ is the predicted class and the true class respectively, $\hat{p}^{z=\hat{y}}$ denotes the the predicted probability of the predicted class. The hyperparameter $\alpha > 0$ plays a role as a learning rate to update the soft label $\widetilde{p}^{z=y}_{r, t}$ based on the difference between the predicted confidence and accuracy in the validation set. Intuitively, if we assign a large $\widetilde{p}^{z=y}_{r}$ to training data, then the network tends to make a high confident prediction and vice versa. Therefore, if the confidence is greater than the accuracy  ($\mathrm{conf}(S^{val}_r) > \mathrm{acc}(S^{val}_r)$) in the validation set, we should reduce $\widetilde{p}^{z=y}_{r}$ to teach the network to be less confident. Otherwise, we should increase $\widetilde{p}^{z=y}_{r}$. In addition, we also need to constrain $\widetilde{p}^{z=y}_{r}$ to be within ($\frac{1}{Z}$, 1] after each update as it stands for the true probability of the correct class, where $Z$ is the number of classes in the dataset.

For a given $\widetilde{p}^{z=y}_{r}$, we can easily obtain $\epsilon_r$ by reversing Eqn. (\ref{eqn:ysoft}),
\begin{equation}\label{eqn:epsilon}
    \epsilon_r = (\widetilde{p}^{z=y}_{r} - 1) \cdot \frac{Z}{1 - Z},
\end{equation}
and the soft labels for all the classes $\widetilde{p}_{r}$ can be computed according to Eqn. (\ref{eqn:y}). We update the soft labels after each training epoch in our experiments.

Note that this adaptive learning mechanism can be easily applied to standard label smoothing without adversarial robustness slicing ($R=1$). In this case, we can replace sweeping the hyperparameter $\epsilon$ with this adaptive learning method, named as ``\textbf{Ada}ptive \textbf{L}abel \textbf{S}moothing'' (\textbf{\ada}). Our proposed \ada and \ar do not increase the inference time: we test \ada and \ar exactly the same as a vanilla model.

\section{Experiments}
\paragraph{Datasets}
We test our method on three datasets CIFAR-10, CIFAR-100 and ImageNet. In addition, we also report performance on the shifted datasets: CIFAR-10-C, CIFAR-100-C and ImageNet-C~\citep{Hendrycks2019BenchmarkingNN}, where there are different types of corruptions (19 types for CIFAR-10, 17 types for CIFAR-100 and 15 types for ImageNet), e.g., noise, blur, weather and digital categories that are frequently encountered in natural images. Each type of corruption has five levels of shift intensity, with higher levels having more corruption. 

\begin{figure}[!t]
     \centering
    \includegraphics[width=0.245\linewidth]{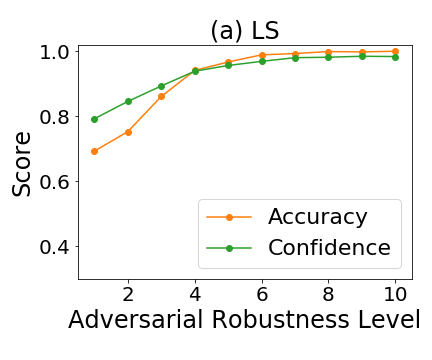}
      \includegraphics[width=0.245\linewidth]{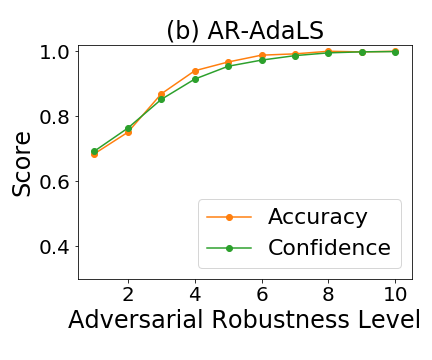}
      \includegraphics[width=0.245\linewidth]{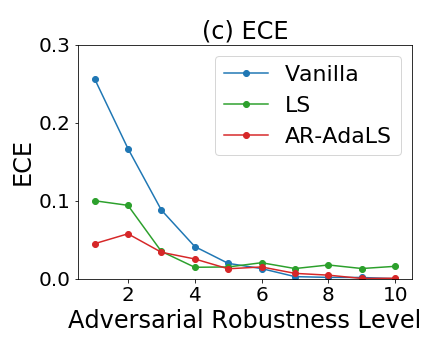}
      \includegraphics[width=0.244\linewidth]{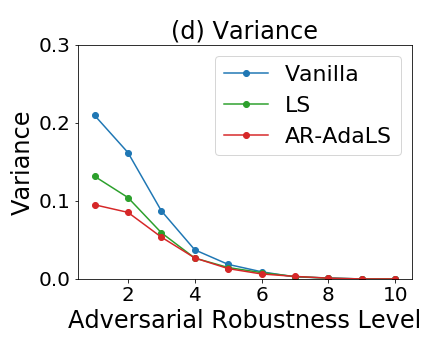}
      \vspace{-6mm}
     \caption{Comparison between LS and our \ar on the clean test set of CIFAR-10. \textbf{(a)} and \textbf{(b):} Accuracy and confidence score of the predicted class in each adversarial robustness subset. \textbf{(c)} and \textbf{(d)}: ECE and variance score of Vanilla, LS and \ar.}
  \label{fig:compare_single}
\end{figure} 
\subsection{How does AR-AdaLS work?}
To have a deeper understanding of how \ar works, in Figure~\ref{fig:compare_single} we visualize the effect of label smoothing (LS) and our \ar. Comparing Figure~\ref{fig:compare_single} (a) and (b), \ar is better at calibrating the data 
than label smoothing, especially on the adversarially unrobust examples (lower adversarial robustness level). Further, we show plots of ECE and variance in Figure~\ref{fig:compare_single} (c) and (d). Both label smoothing and \ar improve model calibration and stability over vanilla model and \ar has the best performance among three methods. This suggests that \ar is better at improving calibration and stability in adversarially unrobust regions, not just on average.

\begin{table}[]
\small
    \centering
    \caption{
    ECE ($\times10^{-2}$) on CIFAR-10 and CIFAR-100. \ar improves calibration and is only rivaled by domain-knowledge based data augmentation or larger ensemble models on CIFAR-100. Adversarial robustness is generated on-the-fly for \ar. All the single-model based results are generated over four independent runs with random initialization. 
    }
    \label{tab:ece-sota}
    \begin{tabular}{c|cc|c|cc}
    \toprule
     Method    &  CIFAR-10 & CIFAR-100 &  Method    &  CIFAR-10 & CIFAR-100 \\ \midrule
      \multicolumn{3}{c|}{Single-model based} & \multicolumn{3}{c}{Data-augmentation based}\\ \midrule
      Vanilla   &  2.49$\pm$0.10& 6.11$\pm$0.24  &   mixup & 0.78$\pm$0.20 & \textbf{1.69$\pm$0.08}\\
      Temperature Scaling & {0.80}$\pm$0.05 & 4.26$\pm$0.07 &  CCAT & 2.37$\pm$0.07 & 7.95$\pm$0.35 \\ \cline{4-6}
      Label Smoothing & 1.07$\pm$0.09& 2.76$\pm$0.26 & \multicolumn{3}{c}{Ensemble based}\\ \cline{4-6}
      \ada & 1.23$\pm$0.02& 2.65$\pm$0.31 &  Mix-n-Match & 0.97 & 2.80 \\
      \ar & \textbf{0.64$\pm$0.02} & \textbf{2.27$\pm$0.16} &  Ensemble of Vanilla & 0.90 &\textbf{2.21}\\
      
      \bottomrule
    \end{tabular}
    \vspace{-5mm}
\end{table}

\vspace{-2mm}
\subsection{AR-AdaLS improves calibration}
\vspace{-1mm}
\paragraph{Baselines}
We compare our proposed \ar with the following 8 different methods: (1) Vanilla model trained with one-hot labels, (2) Temperature Scaling~\citep{Guo2017OnCO}, a post-hoc calibration method where the predicted logits are divided by a temperature which is tuned on the hold-out validation set, (3) label smoothing ({LS})~\citep{Szegedy2016RethinkingTI} that softs labels by sweeping the hyperparameter $\epsilon$ (the smoothing degree) in a range to find the best hyperparameter $\epsilon$, (4) Adaptive Label Smoothing (\textbf{\ada}): we use our proposed adaptive learning mechanism introduced in Section~\ref{sec:adaptive} to automatically learn the hyperparameter $\epsilon$ rather than sweeping to find the best $\epsilon$.  (5) mixup, which is a data augmentation technique originally proposed in~\citep{Zhang2018mixupBE}
and recently found to be able to improve calibration in~\citep{mixup}, (6) Confidence-calibrated adversarial training (CCAT)~\citep{ccat}, a method builds on adversarial training by reducing the confidence in the labels of adversarial examples. Note that there is a significant difference between our \ar and CCAT: CCAT trains a model on the generated adversarial examples to improve a model's adversarial robustness. In contrast, our \ar, trained on the clean training data, is proposed to use the correlation between adversarial robustness and calibration to improve a model calibration performance. (7) ``Ensemble of Vanilla''~\citep{ensemble}, an ensemble of $M$ vanilla models independently trained with random initialization. (8) Mix-n-Match~\citep{mix-n-match}, an ensemble and compositional method proposed for calibration.
All the methods are trained with the same network architecture, i.e., WRN-28-10~\citep{Zagoruyko2016WideRN} on both CIFAR-10 and CIFAR-100, and the same training hyperparameters: e.g., learning rate, batch size, number of training epochs, for fair comparison\footnote{The result of Mix-n-Match in Table~\ref{tab:ece-sota} is from Table 1 reported in the original work~\citep{mix-n-match}, which is trained with the same network architecture, WRN-28-10.}. Please refer to Appendix A for all the training details and hyperparameters.

\vspace{-2mm}
\paragraph{Results}
The expected calibration error of all the methods on CIFAR-10 and CIFAR-100 are displayed in Table~\ref{tab:ece-sota}. We can clearly see that by differentiating the training data based on their adversarial robustness, \ar effectively reduces the calibration error compared to other single-model based methods without significant change in accuracy (see Figure~\ref{fig:ensemble_all} in Appendix) and it is only rivaled by mixup on CIFAR-100, which uses extra domain knowledge through data augmentation. Note that \ar is only trained on the clean training data without any data augmentation compared to mixup~\citep{mixup} and CCAT~\citep{ccat}.

\begin{table*}[t!]

\centering
\vspace{-4mm}
\small
\caption{Mean of ECE ($\times 10^{-2}$) across 19 types of shift for CIFAR-10-C and 15 types of shift for ImageNet-C. Smaller is better. ResNet-29 is used for CIFAR-10 and ResNet-101 is used for ImageNet. The standard deviation of five independent runs for each single model is reported. The best single model
and ensemble model in each shift intensity is highlighted in \textbf{bold}. 
}\label{tab:ece}
\begin{tabular}{c|cc||c|cc}
\toprule
\multicolumn{3}{c||}{Single-model based}  &\multicolumn{3}{c}{Ensemble-based}  \\ \midrule
Methods         & {CIFAR-10-C}  & {ImageNet-C} & Methods & {CIFAR-10-C}  & {ImageNet-C}  \\ \midrule 
Vanilla         &16.7$\pm$0.5 & 10.7$\pm$0.5  & Ensemble of Vanilla  & 6.5&4.2\\ 
LS              & 10.1$\pm$0.4& 8.1$\pm$0.4 &Ensemble of LS  & 4.6 &4.7\\ 
\ada           &9.6$\pm$0.5 & 8.0$\pm$0.2 & Ensemble of \ada  & 5.2 & 4.8\\  \midrule
\ar       & \textbf{6.4}$\pm$0.6 & \textbf{ 6.9$\pm$0.2} & \ensemar & 5.5 & 5.1\\
&&& \arensem & \textbf{4.4} & \textbf{4.0}\\
\bottomrule
\end{tabular}
\vspace{-2mm}
\end{table*}

\subsection{Improve calibration on shifted dataset}
Table~\ref{tab:ece} summarizes the mean calibration error (ECE) on the corrupted datasets: CIFAR-10-C and ImageNet-C~\citep{Hendrycks2019BenchmarkingNN}. 
Looking at all the \emph{single-model} based methods, we can see that \ar significantly outperforms other single-model based methods with the lowest ECE. Contrasting with LS and \ada, we see \ar benefits greatly from the adversarial robustness slicing. As a result, our model learns to give smaller soft labels of the correct class to those adversarially unrobust training data, which can also be considered as outliers of the underlying data distribution~\citep{Carlini2019DistributionDT}. Therefore, when tested on the shifted data that deep networks have been shown to produce pathologically over-confident predictions~\citep{Hendrycks2019BenchmarkingNN}, our model correctly learns to make a relatively lower-confidence prediction, resulting in a better calibration performance. 

In addition, we also compare \ar with ``Ensemble of Vanilla''~\citep{ensemble}, which is shown to be the best model for models' calibration under distributional shift~\citep{snoek2019can}.  
The result of Ensemble of Vanilla is an ensemble of $M=5$ vanilla models independently trained with random initialization. 
We can see that \ar achieves comparable calibration performance on CIFAR-10 and the ensemble is better under highly shifted data on ImageNet. 
\paragraph{Combination with deep ensembles} We further discuss the following two ways to combine \ar with ensembles:
\begin{itemize}[leftmargin=20pt]
    \item \textbf{\ensemar:} As in~\citet{ensemble}, we ensemble \ar by training multiple independent \ar models with random initialization, and average their predictions at inference. 
   \item \textbf{\arensem:} 
Instead of computing soft labels independently for each \ar, we perform \ar on the ensembled predictions, i.e., in Eqn (\ref{eqn:softlabel}) we compute confidence and accuracy based on the average of $M=5$ model predictions. Each model is then supervised with the same soft labels. 
We will see this slight distinction in  training is quite important. 
\end{itemize}

As shown in Table~\ref{tab:ece}, naively combining deep ensembles with \ar (\ensemar) could not effectively improve models' calibration (see more details in Appendix~\ref{sec:ens}). In contrast, \arensem, which adaptively adjusts smoothing to keep the ensemble models well calibrated, performs the best under distributional shift on both CIFAR-10 and ImageNet. 

\subsection{Improve model stability} 
\begin{figure*}[t!]
     \centering
     \includegraphics[width=0.95\linewidth]{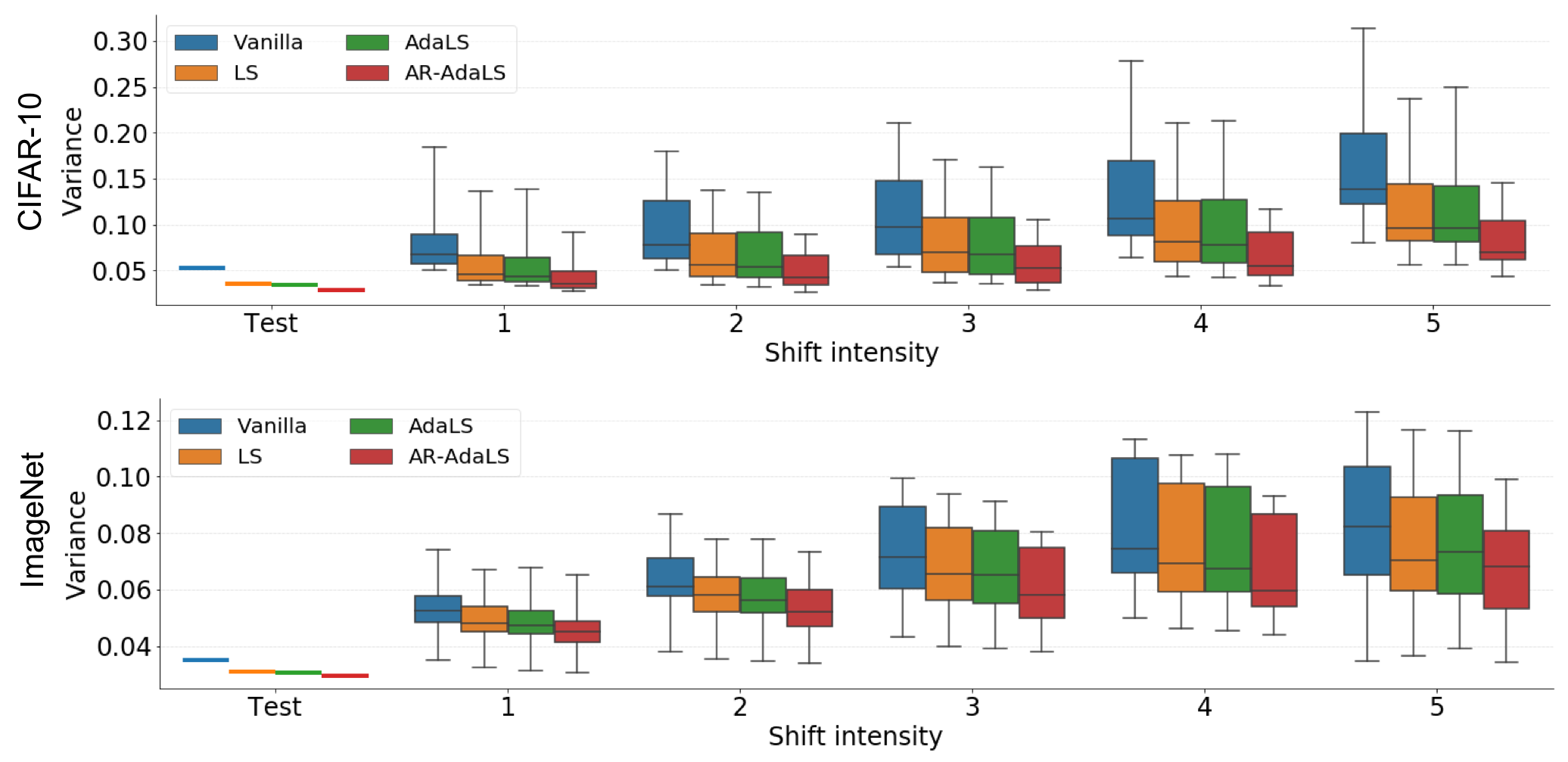}
    \vspace{-2mm}
     \caption{Variance on clean test and shifted data on CIFAR-10 and ImageNet. For each shift intensity, we show the results with a box plot summarizing the 25th, 50th, 75th quartiles across 19 shift types on CIFAR10-C and 15 shift types on ImageNet-C. The error bars indicate the $min$ and $max$ value across different shift types. ResNet-29 is used for CIFAR-10 and ResNet-101 is used for ImageNet.}
  \label{fig:var}
  \vspace{-2mm}
\end{figure*}

Since we observe in Figure~\ref{fig:cor} that the most adversarially unrobust data points also have very unstable predictions, we test \ar to see if it can help improve model stability, which is of great value in practice where high variance of a model is bad for churn \citep{stability_gupta}. In Figure~\ref{fig:var} we can see that \ar can effectively reduce the variance of a model compared to a vanilla model and label smoothing on CIFAR-10 and ImageNet. Please refer to Table~\ref{tab:var} in Appendix for numerical numbers on both datasets.

\subsection{Improvements on out-of-distribution data} 
\begin{figure}[t!]
     \centering
    \includegraphics[width=0.3\linewidth]{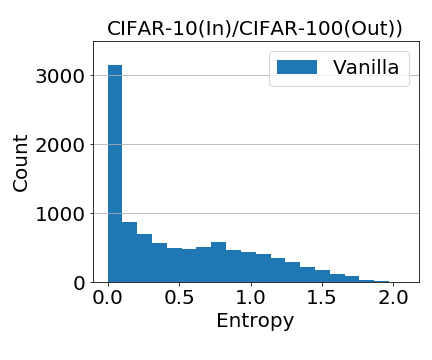}
      \includegraphics[width=0.3\linewidth]{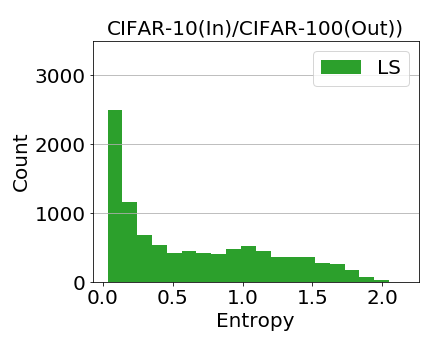}
      \includegraphics[width=0.3\linewidth]{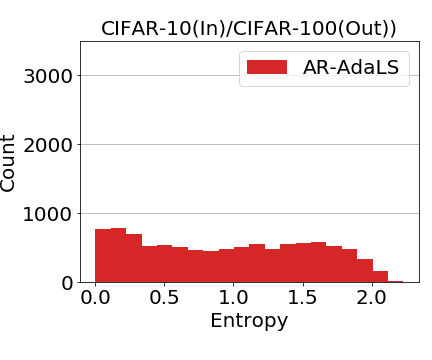}
      \vspace{-3mm}
     \caption{Histogram of predictive entropy on out-of-distribution data. Each model is trained on CIFAR-10 and tested on CIFAR-100. The network architecture is WRN-28-10 and adversarial robustness in \ar is generated on-the-fly.}
  \label{fig:ood}
\end{figure}

We further study the performance of \ar when predicting on out-of-distribution (OOD) data. Following~\citep{snoek2019can}, we compare the performance of Vanilla, Label Smoothing and \ar by plotting the histogram of the entropy on the OOD data (higher entropy on OOD is better). As shown in Figure~\ref{fig:ood}, each model is trained on CIFAR-10 dataset and then tested on CIFAR-100 dataset. We can clearly see that \ar significantly reduces the number of low-entropy predictions on OOD data. In addition, using CIFAR-10/CIFAR-100 as in-distribution/out-of-distribution data, we also report the Area under the ROC curve (AUROC) of label smoothing, mixup and \ar. The AUROC score of standard label smoothing and mixup is 0.832$\pm$0.005 and 0.821$\pm$0.003 respectively, whereas our \ar achieves 0.885$\pm$0.003. This demonstrates the effectiveness of \ar even on fully out-of-distribution data.

\begin{figure}[t]
     \centering

      \includegraphics[width=0.65\linewidth]{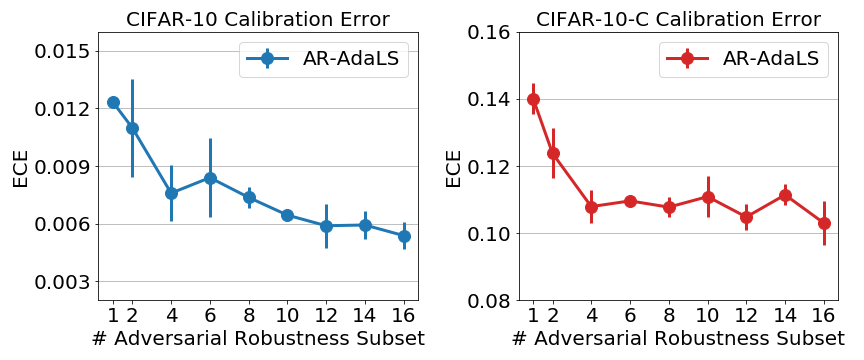}
      \vspace{-2mm}
     \caption{ECE on CIFAR-10 and CIFAR-10-C of \ar with varying number of adversarial robustness subset $R$. Note that when $R=1$, \ar becomes \ada. The results are based on 4 independent runs.}
  \label{fig:sensitivity}
  \vspace{-2mm}
\end{figure} 
\begin{table*}[t]
\small
\centering
\caption{Ablation study of \ar on CIFAR-100 and CIFAR100-C (corrupted).  We report both accuracy ($\times 10^{-2}$) and expected calibration error ($\times 10^{-2}$), denoted by \textbf{Acc} and \textbf{ECE} for the clean test set, and \textbf{cAcc} and \textbf{cECE} for CIFAR100-C. Arrow indicates the better direction; best calibration is \textbf{bolded}.}\label{tab:cifar100}
\begin{tabular*}{0.95\textwidth}{l @{\extracolsep{\fill}} ccccc}
\toprule
\textbf{Method}   & {Vanilla} & {\begin{tabular}[c]{@{}c@{}}Label\\ Smoothing\end{tabular}} & {\begin{tabular}[c]{@{}c@{}}Temperature\\ Scaling\end{tabular}} &  {\begin{tabular}[c]{@{}c@{}}\ar\\ (pre-compute)\end{tabular}} & {\begin{tabular}[c]{@{}c@{}}\ar\\ (on-the-fly)\end{tabular}} \\
\midrule
\textbf{Acc/cAcc} ($\uparrow$) & 79.2/52.0        & 78.9/51.7   & 79.2/52.0   & 79.3/52.2                                                                  & 79.2/52.1                                                   \\
\textbf{ECE/cECE} ($\downarrow$) & 6.1/18.2         & 2.8/16.3    & 4.3/14.0    & 2.6/14.2                                                                   & \textbf{2.3/13.2}  \\\bottomrule                                                     
\end{tabular*}
\end{table*}
\subsection{Sensitivity analysis}\label{sec:ablation}
\paragraph{Sensitivity to the number of adversarial robustness subsets} We perform a sensitivity analysis for the number of adversarial robustness subsets $R$. Specifically, we plot the calibration error of \ar with a varying $R$ on the clean CIFAR-10 and corrupted CIFAR-10-C in Figure~\ref{fig:sensitivity}. We can see that there is a significant drop in calibration error (ECE) when we increase the number of adversarial robustness subsets $R$ from 1, where $R=1$ denotes \ada. Further, the calibration error is relatively stable when $R$ is chosen within the range [10, 16]. Thus, we choose $R=10$ for all results shown in this paper for \ar.

\paragraph{Sensitivity to the exactness of adversarial robustness}
To investigate this, we study the performance of \ar using adversarial robustness generated via two different ways: One is ``on-the-fly'': we keep creating adversarial attacks during training, which provides a more precise adversarial robustness ranking but at the cost of great computing time. The other is to ``pre-compute'' adversarial robustness by attacking a vanilla model that is trained with one-hot labels. This is more efficient but at the sacrifice of the precision of adversarial robustness ranking. We perform experiments on CIFAR-100 as an example to compare the performance of \ar based on the adversarial robustness that is ``pre-computed'' or  ``on-the-fly''. As shown in Table~\ref{tab:cifar100}, generating adversarial robustness ``on-the-fly'' can further help improve the calibration performance for \ar on both clean and shifted datasets, compared to pre-computing adversarial robustness. 
Similar patterns are observed on CIFAR-10.\footnote{We did not run on-the-fly \ar for ImageNet due to the computational intensity.} 

Therefore, we can conclude that 1) the exactness of adversarial robustness is helpful for \ar, that is, more precise adversarial robustness leads to a better performance. 2) \ar with an approximation of adversarial robustness (pre-computed) can already significantly improve label smoothing. 
Hence, all results in this paper related to ``\ar'' without further specification are based on pre-computed adversarial robustness for efficiency. This is because our main target is to show that the idea of differentiating the training data based on their adversarial robustness is promising to improve model calibration rather than pushing the results to the best. 

\section{Conclusion}
In this paper, we have explored the correlations between adversarial robustness and calibration. We find across three datasets that adversarially unrobust data points, where small adversarial perturbations to the input are able to fool the classifier into wrong predictions, are more likely to have poorly calibrated and unstable predictions. Based on this insight, we propose \ar to adaptively smooth the labels of the training data based on their adversarial robustness. 
In our experiments we see that \ar is more effective than previous label smoothing methods in improving calibration, particularly for shifted data, and can offer improvements on top of already strong ensembling methods.
We believe this is an exciting new use for adversarial robustness as a means to more generally improve model trustworthiness, not just by limiting adversarial attacks but also improving calibration and stability on unexpected data.  
We hope this spurs further work at the intersection of these areas of research.

\bibliographystyle{icml2021}

\appendix

\section{Implementation Details}\label{sec:implement}
\subsection{CIFAR-10}
\paragraph{ResNet-29} For all the experimental results on ResNet-29 v2~\citep{resnetv2}, we use a batch size of 256. The network is trained with Adam optimizer~\citep{kingma2015variational} for 200 epochs. The initial learning rate is $10^{-3}$ and decayed down to $10^{-4}$ after 80 epochs, $10^{-5}$ after 120 epochs, $10^{-6}$ after 160 epochs and $0.5 \times 10^{-6}$ after 180 epochs. We adapted the following data augmentation and training script at \url{https://keras.io/examples/cifar10_resnet/}. The training mechanism is the same for all the methods that we compare in the main paper. We randomly split the training dataset into training data of 45000 images and 5000 images as the validation set. The test set has 10000 images.

For label smoothing (LS), we sweep the hyperparameter $\epsilon$ within the range [0, 0.1] with a step size 0.01 and find that the network has the best calibration performance on the validation set when $\epsilon = 0.02$.

For Adaptive Label Smoothing (\ada), there is a hyperparameter $\alpha$ which plays a role as learning rate in the adaptive learning mechanism. We choose hyperparameter $\alpha$ based on the calibration performance on the validation set. Specifically, we run experiments with $\alpha \in \{0.005, 0.01, 0.05, 0.1\}$ and find that $\alpha = 0.05$ achieve the best calibration performance.

Similarly, for Adversarial Robustness based Adaptive Label Smoothing (\ar), we choose the hyperparameter $\alpha$ from the set $\{0.005, 0.01, 0.05\}$ and empirically set $\alpha = 0.005$ which has the best calibration performance on the validation set. We update the training labels after each epoch for all the experiments related to \ar, including the experiments on CIFAR-100 and ImageNet. We use the same hyperparameter $\alpha =0.005$ without further tuning for \ar of Ensemble.

All the results of ensemble models are obtained via training 5 independent models with random initializations. 

\paragraph{WRN-28-10} We train a Wide ResNet-28-10 v2~\citep{Zagoruyko2016WideRN} to obtain the state-of-the-art accuracy for CIFAR-10 (e.g., Table 2 in the main text). We adapt the same training details and data augmentation at \url{https://github.com/google/edward2/blob/master/baselines/cifar/deterministic.py}. 

For label smoothing, we sweep the hyperparameter $\epsilon$ within the range [0, 0.1] with a step size 0.01 and find that the network has the best calibration performance on the validation set when $\epsilon = 0.02$. 

For \ada and \ar, the hyperparameter $\alpha$ is set to be 0.005. For \ar that generated with on-the-fly adversarial examples, we recompute the adversarial robustness for training and validation sets after 65, 130 epochs.

For mixup~\citep{Zhang2018mixupBE, mixup},  the mixing
parameter of two images is randomly sampled from a Beta distribution
Beta($\beta, \beta$) at each training iteration. We set $\beta=0.2$ for best calibration performance on in-distribution data.

For CCAT~\citep{ccat}, we observe that training models with adversarial examples bounded with smaller $\ell_\infty$ norm,
e.g., $||\delta||_{\infty} \leq 0.01 $, can benefit more to the calibration with a small accuracy sacrifice
on the clean data. Therefore, we train CCAT with PGD attacks bounded by $||\delta||_{\infty} \leq 0.01 $. The step size and total iterations to generate PGD attacks is $0.025$ and 10 respectively during training. 

All the results of ensemble models on WRN-28-10 are obtained via training 4 independent models with random initializations. 
\subsection{CIFAR-100}
We train a Wide ResNet-28-10 v2~\citep{Zagoruyko2016WideRN} to obtain the state-of-the-art accuracy for CIFAR-100. We adapt the same training details and data augmentation at \url{https://github.com/google/edward2/blob/master/baselines/cifar/deterministic.py}. 

For label smoothing, we e sweep the hyperparameter $\epsilon$ within the range [0, 0.1] with a step size 0.01 and find that the network has the best calibration performance on the validation set when $\epsilon = 0.07$. 

All the hyperparameters used for \ada, \ar, mixup~\citep{Zhang2018mixupBE, mixup} and CCAT~\citep{ccat} are the same as those for CIFAR-10 with WRN-28-10.

All the results of ensemble models on WRN-28-10 are obtained via training 4 independent models with random initializations. 
\subsection{ImageNet}
All the experiments on ImageNet were obtained via training a ResNet-101 v1~\citep{resnetv1} following the training script at \url{https://github.com/google/edward2/blob/master/baselines/imagenet/deterministic.py}. The network is trained with a batch size of 128 for each TPU core with SGD optimizer for 90 epochs. The input image is normalized (divided by 255) to be within [0,1]. We randomly divide 50000 validation images into validation set with 25000 images and test set with 25000 images. Note that the same dataset and training mechanisms are used for all the methods that we compare in the main paper.

For Label Smoothing (LS), we sweep the hyperparameter $\epsilon$ within the range [0, 0.1] with a step size 0.01 and find that the best calibration performance on the validation set is achieved by setting $\epsilon=0.02$.

For Adaptive Label Smoothing (\ada), we sweep the hyperparameter $\alpha$ in the set 
$\{0.005, 0.01, 0.03, 0.05, 0.1\}$ and set it to be $\alpha=0.03$ for the best calibration performance on the validation set.

We empirically set $\alpha=0.001$ for \ar in the first 60 epochs of the training and then increase it to $0.05$ for the next 30 epochs. The same hyperparameter $\alpha$ is used for \ar of Ensemble without further tuning.

All the ensemble models are a combination of 5 independent models with random initializations.

\subsection{CW attacks}
To compute the adversarial robustness, we construct $\ell_2$ based CW attacks~\citep{carlini2017towards} following the code at \url{https://github.com/tensorflow/cleverhans/blob/master/cleverhans/attacks/carlini_wagner_l2.py}. Specifically, we set the binary search steps to be 3, max iterations to be 500 and learning rate to be $0.005$. The generated untargeted CW attacks can achieve 100$\%$ success rate for all the datasets that we consider: CIFAR-10, CIFAR-100 and ImageNet. We set the number of adversarial robustness training subset and validation subset to be $R=10$ respectively. 

\begin{figure*}[ht]
     \centering
     \includegraphics[width=0.98\linewidth]{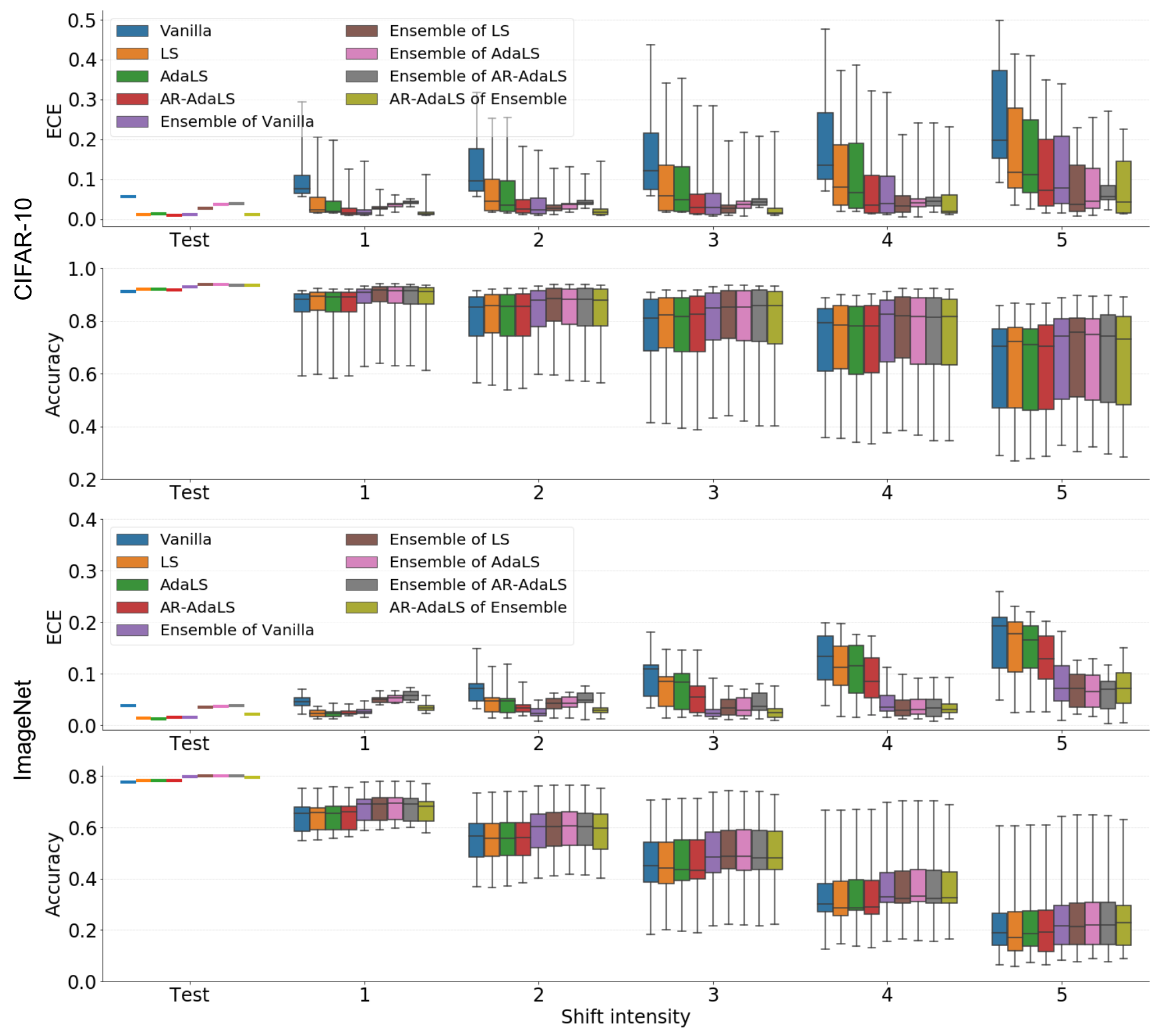}\\
     \caption{Comparison of ensemble models: ECE and Accuracy on both clean test data and shifted data on CIFAR-10 and ImageNet. For each intensity of shift, we show the results with a box plot summarizing the 25th, 50th, 75th quartiles across 19 types of shift on CIFAR-10-C and 15 types of shift on ImageNet-C. The error bars indicate the $min$ and $max$ value across different shift types.}
  \label{fig:ensemble_all}
\end{figure*} 

\begin{figure}[t!]
     \centering
    \includegraphics[width=0.24\linewidth]{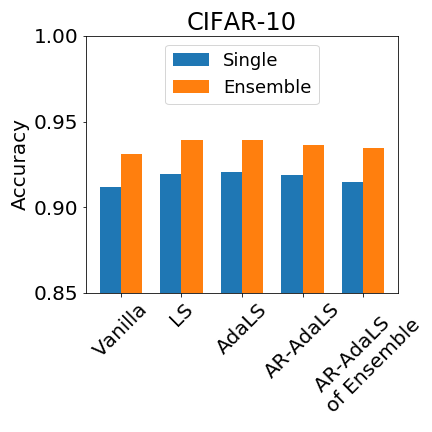}
      \includegraphics[width=0.24\linewidth]{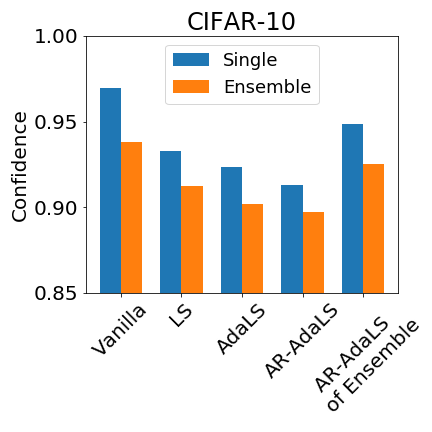}
      \includegraphics[width=0.24\linewidth]{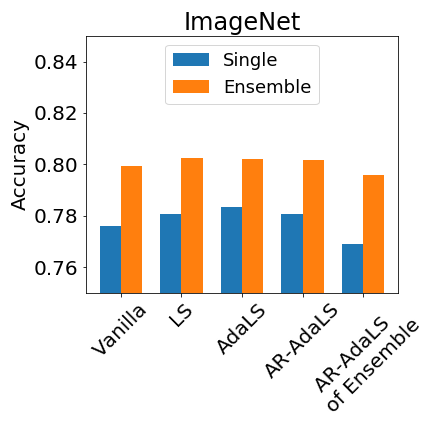}
      \includegraphics[width=0.24\linewidth]{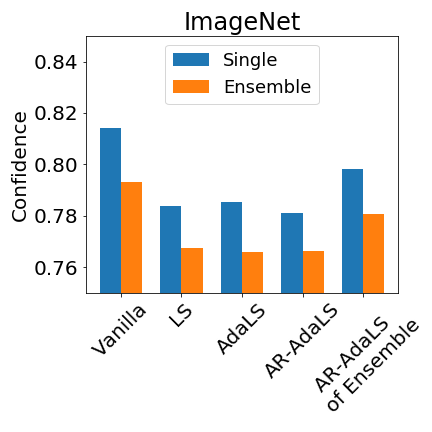}
      \vspace{-4mm}
     \caption{Comparing accuracy and confidence of the predicted class between single model and the corresponding ensemble model for each method.}
  \label{fig:compare_ensemble}
\end{figure}

\begin{figure*}[t!]
     \centering
     \includegraphics[width=0.19\linewidth]{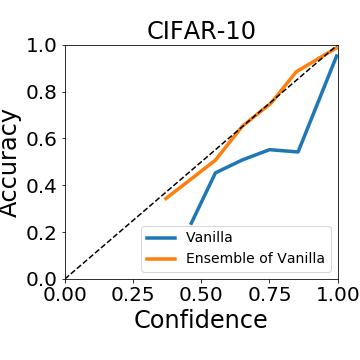}
    \includegraphics[width=0.19\linewidth]{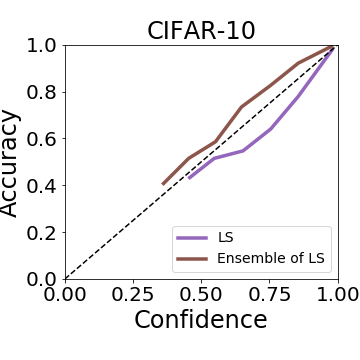}
      \includegraphics[width=0.19\linewidth]{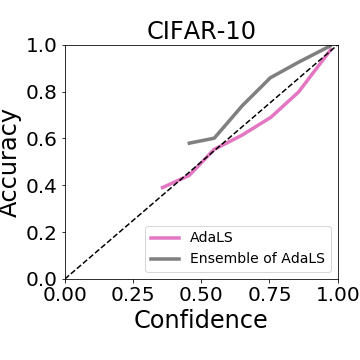}
      \includegraphics[width=0.19\linewidth]{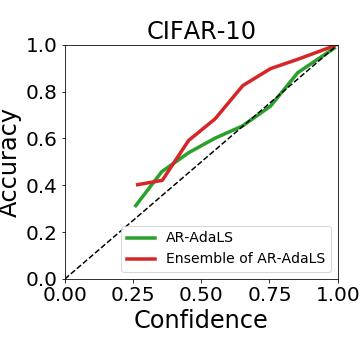}
      \includegraphics[width=0.19\linewidth]{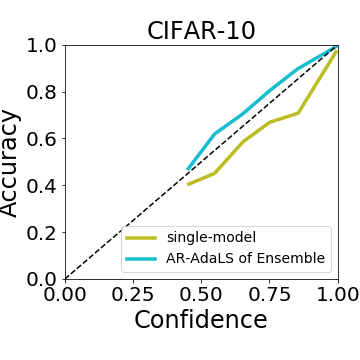}\\
      \includegraphics[width=0.19\linewidth]{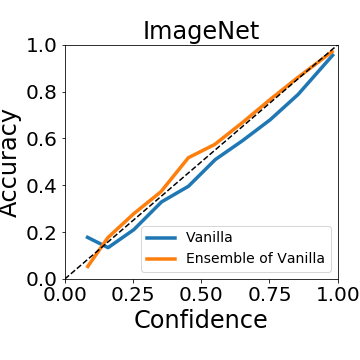}
    \includegraphics[width=0.19\linewidth]{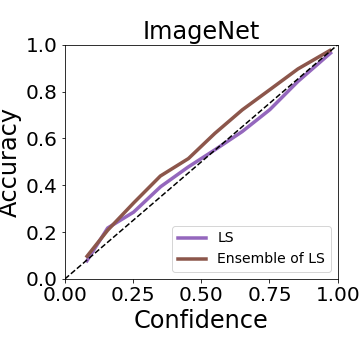}
      \includegraphics[width=0.19\linewidth]{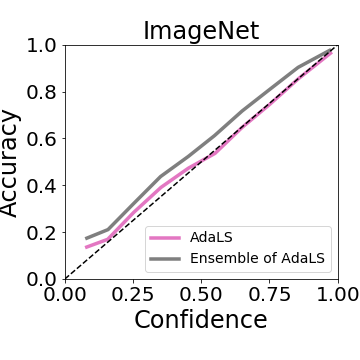}
      \includegraphics[width=0.19\linewidth]{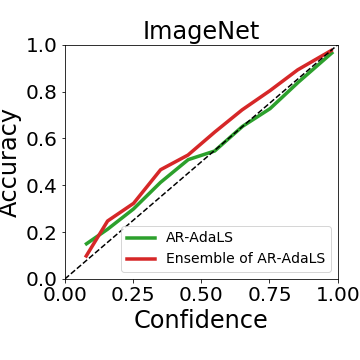}
      \includegraphics[width=0.19\linewidth]{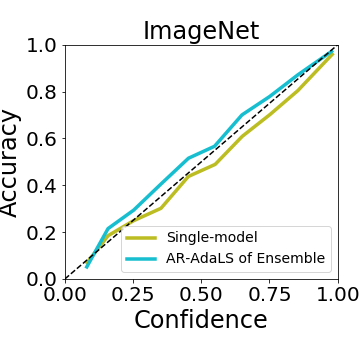}
      \vspace{-4mm}
     \caption{Reliability diagram of accuracy versus confidence of single model and ensemble model on the clean test of CIFAR-10 and ImageNet. The perfect calibrated model should be aligned with the diagonal dotted line (above is under-confident, below is over-confident).}
  \label{fig:compare_all}
\end{figure*}  

\section{Discussion of AR-AdaLS Combined with Ensembles}\label{sec:ens}
In Figure~\ref{fig:ensemble_all}, we show ECE and accuracy of all the single-models and their corresponding ensembles on the clean test and shifted CIFAR-10 and ImageNet. At a high level, we see that all the ensemble models that we compare have similar accuracy, which are higher than single-models. \arensem performs the best across both clean test data and all intensities of shifted data 
in terms of calibration. 

Looking more closely, some trends emerge:
all of the ensemble methods perform relatively well for highly shifted data (intensity 4--5), but Ensemble of LS, Ensemble of \ada, \ensemar perform much worse on less shifted and clean test data. Digging deeper, we display the confidence of the predicted class and accuracy of each single model and the corresponding ensemble models on the clean test set of CIFAR-10 and ImageNet in Figure~\ref{fig:compare_ensemble}. We can clearly see that the ensemble models generally increase accuracy and decrease confidence compared to a single model, which results from the disagreement of the prediction of each single model in ensembles. Therefore, naive deep ensembles can improve calibration on highly shifted data where single-model is over-confident 
but can harm calibration if applied to a well-calibrated single-model. This is made clearer in Figure \ref{fig:compare_all}: while deep ensembles make over-confident vanilla model well calibrate, it leads the well calibrated models, e.g., \ar, to be under-confident.
From this perspective, \arensem avoids this issue by adaptively adjusting smoothing to keep the ensembles well calibrated on both clean and shifted dataset.

\begin{table}[t!]
\small
\centering
\caption{Mean of variance ($\times 10^{-2}$) across 19 types of shift for CIFAR-10-C and 15 types of shift for ImageNet-C. Best in \textbf{bold}.}\label{tab:var}
\begin{tabular}{c|ccccc|c||ccccc|c}
\toprule

Dataset         & \multicolumn{6}{c||}{CIFAR10-C}  &  \multicolumn{6}{c}{ImageNet-C}\\ \midrule
 Shift Intensity & 1    & 2    & 3    & 4    & 5  & Mean & 1    & 2    & 3    & 4    & 5  & Mean  \\ \midrule
Vanilla         & 7.85 & 9.69 & 11.2 & 13.1 & 16.0  & 11.6  &  5.28 & 6.39 & 7.37 & 8.23 & 8.29 & 7.11 \\ 
LS              & 5.54 & 6.95 & 8.11 & 9.65 & 11.8 & 8.41              &  4.86 & 5.84 & 6.78 & 7.55 & 7.41& 6.49 \\ 
\ada           & 5.47 & 6.87 & 7.95& 9.44 & 11.5 & 8.25 & 4.79 & 5.77 & 6.66 & 7.51 & 7.56 & 6.46\\ 
\ar        & \textbf{4.21} & \textbf{5.06} & \textbf{5.73} & \textbf{6.66} & \textbf{8.24} & \textbf{5.98}  & \textbf{4.53} & \textbf{5.49} & \textbf{6.12} & \textbf{6.76} & \textbf{6.66} & \textbf{5.91} \\  
\bottomrule
\end{tabular}
\end{table}

\end{document}